\documentclass[letterpaper, 10 pt, conference]{ieeeconf}
\usepackage{graphicx}
\usepackage{amsmath}
\usepackage{amssymb}
\usepackage{threeparttable}
\usepackage{booktabs}
\usepackage{color}
\usepackage{algorithm}
\usepackage{algorithmic}
\usepackage{pifont}
\usepackage{mathrsfs}
\usepackage{multirow}
\usepackage{threeparttable}
\usepackage{flushend}
\let\labelindent\relax\usepackage{enumitem}
\usepackage[colorlinks,linkcolor=blue]{hyperref}
\setlist[itemize]{leftmargin=*}
\newcommand{\cmark}{\ding{51}}
\newcommand{\xmark}{\ding{55}}

\IEEEoverridecommandlockouts
\begin{document}

\title{\LARGE \bf Aligning Knowledge Graph with Visual Perception \\ for Object-goal Navigation}

\author{Nuo Xu,\ Wen Wang,\ Rong Yang,\ Mengjie Qin,\ Zheyuan Lin,\ Wei Song*,\ Chunlong Zhang*,\ Jason Gu,\ Chao Li\
\thanks{Nuo Xu, Wen Wang, Rong Yang, Mengjie Qin, Zheyuan Lin, Wei Song, Chunlong Zhang, Jason Gu and Chao Li are with ZhejiangLab, Hangzhou 311100, China (e-mail:\{nuo.xu, wangwen, yang\_rong, qmj, linzhy, weisong, zcl1515, jgu, lichao\}@zhejianglab.com). Jason Gu is also with Department of Electrical and Computer Engineering, Dalhousie University, Halifax, Canada. Wei Song and Chunlong Zhang are co-corresponding authors (*).}
}

\markboth{2024 IEEE International Conference on Robotics and Automation (ICRA 2024)}%
{Shell \MakeLowercase{\textit{et al.}}: Bare Demo of IEEEtran.cls for IEEE Journals}

\maketitle
\pagestyle{empty}
\thispagestyle{empty}

\begin{abstract}
Object-goal navigation is a challenging task that requires guiding an agent to specific objects based on first-person visual observations. The ability of agent to comprehend its surroundings plays a crucial role in achieving successful object finding. However, existing knowledge-graph-based navigators often rely on discrete categorical one-hot vectors and vote counting strategy to construct graph representation of the scenes, which results in misalignment with visual images. To provide more accurate and coherent scene descriptions and address this misalignment issue, we propose the Aligning Knowledge Graph with Visual Perception (AKGVP) method for object-goal navigation. Technically, our approach introduces continuous modeling of the hierarchical scene architecture and leverages visual-language pre-training to align natural language description with visual perception. The integration of a continuous knowledge graph architecture and multimodal feature alignment empowers the navigator with a remarkable zero-shot navigation capability. We extensively evaluate our method using the AI2-THOR simulator and conduct a series of experiments to demonstrate the effectiveness and efficiency of our navigator. Code available:  \href{https://github.com/nuoxu/AKGVP}{https://github.com/nuoxu/AKGVP}.
\end{abstract}

\IEEEpeerreviewmaketitle

\section{Introduction}
Object-goal navigation \cite{batra2020objectnav,li2022object} presents a formidable challenge as agents endeavor to navigate unfamiliar and dynamic environments based on visual observations to locate specific objects. However, traditional approaches encounter difficulties in locally visible settings due to the absence of spatial layout information, impeding efficient navigation. Achieving successful object-goal navigation necessitates surpassing robust perception and enabling agents to possess a comprehensive understanding of their surroundings, empowering them to make intelligent decisions to accomplish their goals. This calls for a seamless integration of perception, language, and action, where agents adapt their navigational strategies based on real-time visual input, language instructions, and sensory feedback. Successful visual navigation hinges upon mapping visual observations to actions and teaching machines to perceive objects, infer attributes, and reason about relationships. Agents must exhibit enhanced abilities to reason, plan, and execute goal-directed actions that remain adaptable to dynamic environmental changes. Consequently, learning informative visual representations and modeling Semantic Maps \cite{han2021semantic} or Knowledge Graphs \cite{ji2021survey} of the scenes become indispensable for augmenting performance in unfamiliar environments.

\begin{figure}[t!]
	\centering
	\includegraphics[width=0.93\linewidth]{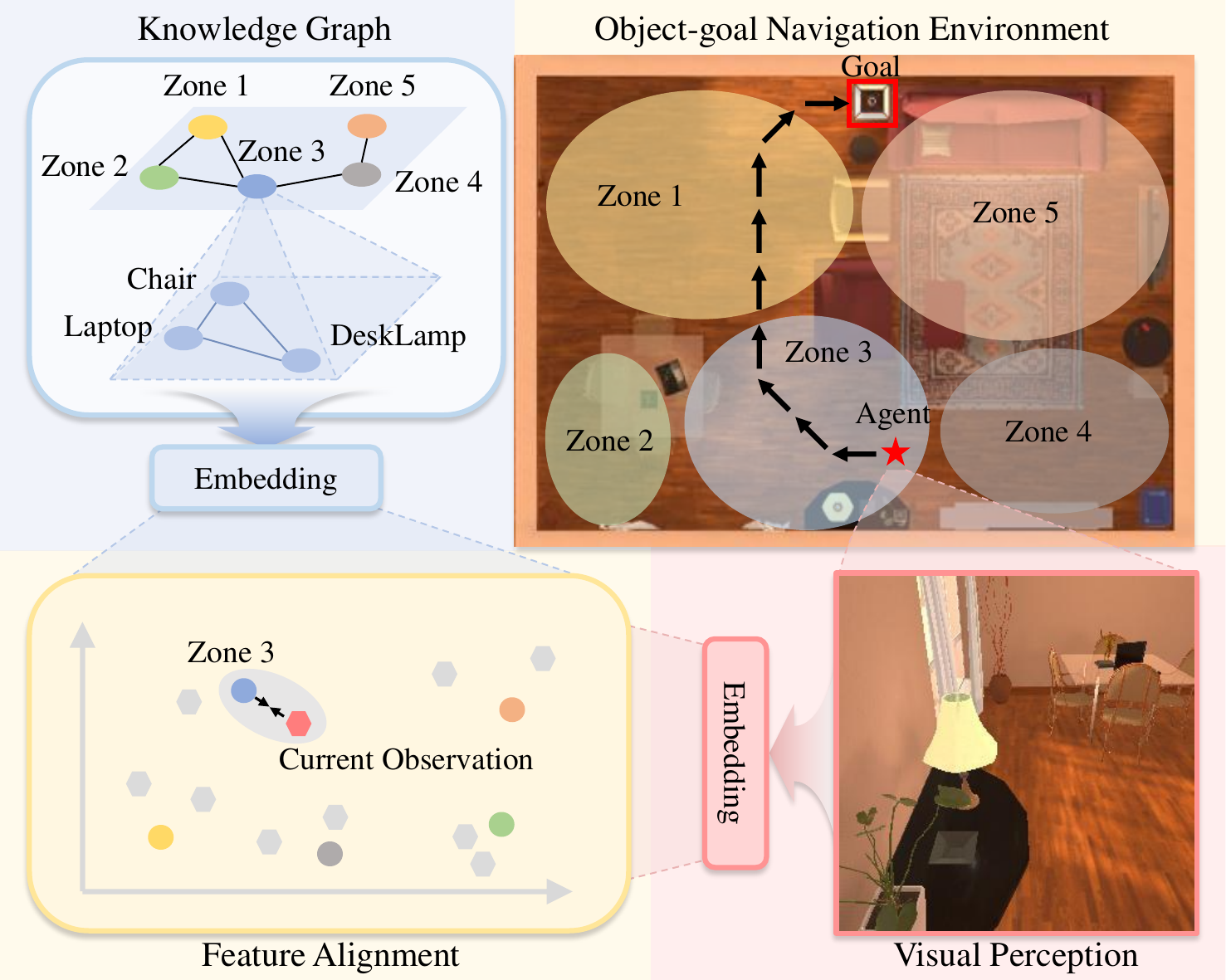}
	\caption{The core idea of our AKGVP method. In order to effectively describe the same environment, we can leverage two modalities of data: knowledge graphs derived from natural language descriptions and observation images characterized by visual descriptions. Our primary objective is to align these two modalities within a shared feature space, facilitated by visual-language pre-training. Ultimately, these modalities are fused for decision-making.}
	\label{fig:idea}
	\vspace{-0.45cm}
\end{figure}

Object-goal navigators encompass three primary categories: end-to-end methods, knowledge-graph-based methods, and semantic-map-based methods. In the end-to-end approach, agents gather observations (including RGB images, depth, and pose) from the environment at each time step. Utilizing these inputs and object categories, agents employ various temporal models, such as RNNs \cite{ye2021auxiliary}, LSTMs \cite{mousavian2019visual,wortsman2019learning,mayo2021visual,du2021vtnet,khandelwal2022simple}, memory-based models \cite{du2020learning}, or even without explicit temporal modeling \cite{zhu2017target,shen2019situational,maksymets2021thda}, to determine the subsequent actions. The end-to-end approach exclusively relies on inputs like RGB images for direct action generation. Nevertheless, it often struggles with generalizing in complex scenarios. To overcome this challenge, researchers have introduced additional modeling techniques, such as knowledge graphs \cite{wu2019bayesian,yang2019visual,zhang2021hierarchical,zhang2023layout} and semantic maps \cite{chaplot2020object,liang2021sscnav,wani2020multion,ramakrishnan2022poni,gadre2023cows}. Knowledge graphs provide robust representation capabilities, capturing entity correlations and contextual information. Graph structures integrate vast amounts of structured and semi-structured data. In contrast, semantic maps enable spatial perception, describing entity positions, directions, and geometry. They offer real-time updates, aiding rapid adaptation in dynamic environments.

Nevertheless, the existing knowledge-graph-based navigators suffer from a significant drawback, primarily stemming from their extensive dependence on discrete categorical vectors and vote-counting strategy for constructing graph representation of the scenes. Unfortunately, this approach frequently leads to a misalignment between these categorical representations and the actual visual images of the scene. Relying solely on decision-making losses, such as reinforcement learning, proves challenging in effectively aligning the two modalities. Addressing this limitation is of paramount importance in order to enhance the performance and accuracy of object-goal navigators in real-world scenarios. With this in mind, our research endeavors to push the boundaries of knowledge-graph-based navigators by focusing on improving their capabilities and overall performance.

To address the aforementioned challenges, we propose the Aligning Knowledge Graph with Visual Perception (AKGVP) for object-goal navigation. As depicted in Fig.\;\ref{fig:idea}, our approach introduces a continuous modeling framework that captures the hierarchical knowledge-graph-based scene architecture, leading to more accurate and coherent scene descriptions. Additionally, we leverage visual-language pre-training techniques to align scene language description with visual perception, bridging the gap between semantic graph and visual understanding. This alignment enhances the comprehension of instructions and scenes for agent within the context of navigation tasks, enabling a more refined representation of the visual environment. Moreover, the integration of a continuous knowledge graph architecture and multimodal feature alignment empowers the navigator with a remarkable zero-shot navigation capability \cite{pourpanah2022review,khandelwal2022simple,gadre2023cows}. In this work, we extensively evaluate the effectiveness and efficiency of our proposed AKGVP method using the AI2-THOR \cite{ai2thor} simulator, a widely adopted platform for embodied navigation research. Through a series of rigorous experiments, we demonstrate the superior performance of AKGVP in achieving object-finding tasks. The contributions of this research are as follows:

\begin{itemize}
  \setlength{\itemsep}{0pt}
  \item A novel continuous knowledge-graph-based scene modeling framework is introduced that captures the intricate hierarchical scene architecture, resulting in more accurate and coherent scene descriptions.
  \item Moreover, one visual-language pre-training technique is leverage advanced to effectively align scene language descriptions with visual perception, proposing our Aligning Knowledge Graph and Visual Perception (AKGVP).
  \item In this study, we extensively evaluate the effectiveness and efficiency of our proposed AKGVP using the AI2-THOR simulator. Through rigorous experiments, the superior performance of AKGVP is demonstrated.
\end{itemize}

\section{Methodology}
\label{sec:objnav}
In the Object-goal Navigation (ObjNav) task, the agent is situated within the 3D indoor environments $Q$ and aims to efficiently approach predefined goal objects $G$ in the fewest steps. During each episode, the agent is randomly initialized at position $l=\{x, z, \theta_1, \theta_2\}$ within the environment $q\in Q$. Here, $(x, z)$ represents the plane coordinates, while $\theta_1$ and $\theta_2$ correspond to the yaw and pitch angles, respectively. At each timestamp $t$, the agent learns a policy function denoted as $\pi(a_t | s_t, g)$, which predicts the appropriate action $a_t\in \mathcal{A}$ based on the current state (first-person observation) $s_t$ and the goal object $g\in G$. Here, $\mathcal{A}=\{$$MoveAhead,$ $RotateLeft,$ $RotateRight,$ $LookDown,$ $LookUp,$ $Done\}$. Notably, the agent independently determines the completion of an action, relying on its own judgment rather than environmental cues. The movement of agent occurs within a discretized scene space with intervals of 0.5m. The rotation and the vertical tilt angle remain fixed at 45 and 30 degrees, respectively. An episode is deemed successful when the agent selects the Done action while the goal object is visible within a threshold distance of 1.5m, indicating proximity. Failure occurs if the agent does not meet this criterion.

\begin{figure*}[t!]
	\centering
	\includegraphics[width=0.73\linewidth]{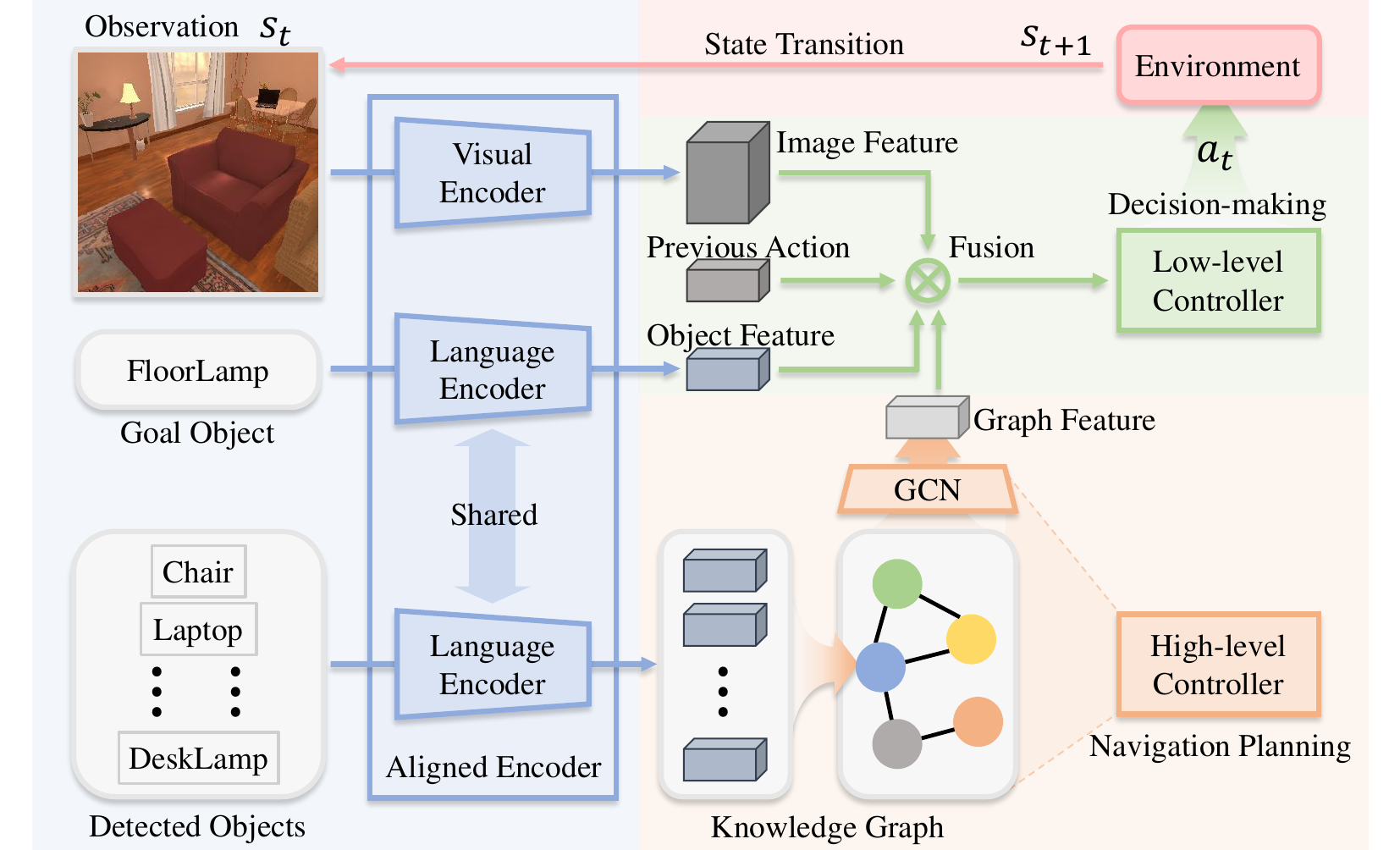}
	\caption{Pipeline of our AKGVP method. AKGVP is composed of three essential components: an aligned encoder, a high-level controller, and a low-level controller. The encoder plays a crucial role by separately encoding images and natural language, facilitating feature alignment through multimodal pre-training. The high-level controller leverages knowledge graph modeling to effectively plan sub-goals for the navigator, directing the movement of agent across different zones. On the other hand, the low-level controller utilizes the fused multimodal information to make informed action decisions, enabling the agent to interact with the environment and control its movements adeptly.}
	\label{fig:pipeline}
	\vspace{-0.35cm}
\end{figure*}

\subsection{Overview}
\label{sec:overview}
As illustrated in Fig.\;\ref{fig:pipeline}, the AKGVP framework consists of three indispensable components: an aligned encoder, a high-level controller, and a low-level controller. The aligned encoder assumes a critical role in separately encoding images and natural language, facilitating the alignment of features through multimodal pre-training. Harnessing knowledge graph modeling, the high-level controller adeptly plans sub-goals for the navigator, effectively guiding the movement of agent across diverse zones. In parallel, the low-level controller leverages the fused multimodal information to make well-informed action decisions, enabling the agent to interact with the environment and exert precise control over its movements. Fig.\;\ref{fig:pipeline} offers a visual depiction of the intricate interplay between these components within the AKGVP framework. In the subsequent sections, we will delve into the detailed exposition of each of these three components.

\subsection{MultiModal Feature Alignment}
\label{sec:aligning}
In this section, we present the aligned encoder, a pivotal component responsible for vision-language alignment and feature extraction.

\noindent \textbf{Multimodal Pre-training.}\; Multimodal pre-training, exemplified by the pioneering CLIP \cite{radford2021learning} model, represents a technique wherein a model is trained on diverse modalities, such as images and natural language, fostering the acquisition of rich representations that capture intermodal correlations. This pre-training paradigm is designed to empower the model with the ability to comprehend and reason about the intricate relationships existing between visual and textual information. Through pre-training, CLIP learns to establish associations between images and text via the juxtaposition of positive pairs (matching image-text pairs) against negative pairs (non-matching pairs). This process effectively encourages the model to align similar images with their corresponding textual descriptions within a shared feature space.

\noindent \textbf{Aligned Encoder.}\; In fact, CLIP (ResNet-50 \cite{he2016deep}) is applied in the Aligned Encoder module within the AKGVP model, comprising two components: the visual encoder and the language encoder. To preserve the spatial position information in the image, the last fully connected layer of the visual encoder is removed, while the language encoder remains intact. During training, the parameters of the aligned encoder are consistently frozen, requiring no updates. As depicted in Fig.\;\ref{fig:pipeline}, the Image Feature $f_{img}$ is represented as a 7$\times$7$\times$2048 tensor, while the Object Feature $f_{obj}$ is a continuous 1024-dimensional vector. These architectural choices contribute to the efficient encoding and representation of visual and textual information in AKGVP. In contrast, conventional knowledge-graph-based navigators \cite{wu2019bayesian,yang2019visual,zhang2021hierarchical,zhang2023layout} frequently employ discrete one-hot vectors to represent objects in the scene, lacking alignment with visual perception. The application of CLIP presents compelling advantages, including the cultivation of robust representations, improved generalization capabilities, and zero-shot ability. These advantages will collectively augment the capacity of the model to navigate towards the desired goal object in a more cognizant and efficient manner,  in the realm of object-goal navigation.

\subsection{Knowledge Graph Modeling}
\label{sec:graph}
The high-level controller leverages Graph Convolutional Networks (GCNs) \cite{wu2020comprehensive} to extract prior information from the knowledge graph. This extracted high-level semantic knowledge serves as a valuable guide for the agent to navigate and explore unknown environments more effectively, using the power of the knowledge graph.

\noindent \textbf{Spatial Location Clustering.}\; In order to construct a comprehensive knowledge graph that captures the object distribution within the training environment $q\in Q$, the agent meticulously explores all positions $l=\{x, z, \theta_1, \theta_2\}$. At each position $(x, z)$, the agent diligently detects the presence of each goal object $g_k\in G$ within its observations by Faster RCNN \cite{ren2015faster}, meticulously recording this crucial information using $f_{(x,z)}$. The detection results from various viewing angles $(\theta_1, \theta_2)$ will be averaged when sharing the same coordinates $(x, z)$.
\begin{equation}
	\label{eq:fpos}
	f_{(x,z)}=\frac{1}{\lvert\theta_1 \rvert \lvert\theta_2 \rvert \sum_{k}{\alpha_k}}\sum\limits_{\theta_1, \theta_2, k} {\alpha_k f_{obj,k}}.
\end{equation}
where $\alpha_k$ signifies the presence of the goal object $g_k$ within the observation field of view. It takes the value of 1 if the object is visible and 0 otherwise. $f_{obj,k}$ denotes the Object Feature associated with the goal object $g_k$. Additionally, $\lvert\theta_1 \rvert$ and $\lvert\theta_2 \rvert$ represent the total choices of yaw and pitch angles, respectively. It is worth noting that scenes with similar characteristics often exhibit a recurring pattern of typical regions adorned with similar object types and layouts. For instance, a bedroom typically encompasses an area comprising a bed, a chest, and pillows, forming a distinctive arrangement. In these analogous scenes, we employ K-Means clustering techniques to group the recorded information $f_{(x,z)}$ corresponding to each spatial coordinate $(x, z)$ across each room. This process effectively divides the room into distinct zones $\{Z_{m}|(x_i,z_i) \in Z_{m}, m=1,\cdots,M\}$, reflecting the underlying variations in object distribution and facilitating more granular analysis of the environment.

\noindent \textbf{Graph Definition.}\; We denote our knowledge graph as $\Omega = (V, E)$, where $V$ represents the nodes and $E$ represents the edges connecting the nodes. In our specific case, \textit{e.g.}, a certain room, each node $v_m\in V$ corresponds to each zone $Z_{m}$ with its cluster center $\delta(v_m)$. Meanwhile, each edge $e(v_m,v_n)\in E$ captures the adjacent probability between two zones ($Z_{m}$, $Z_{n}$). 
\begin{equation}
	\label{eq:node}
	\delta(v_m)=f_{Z_m}=\frac{\sum_{(x_i,z_i) \in Z_{m}}{f_{(x_i,z_i)}}}{\lvert Z_{m} \rvert}.
\end{equation}
\begin{equation}
	\label{eq:edge}
	\begin{split}
	e(v_m,v_n)&=\frac{\sum_{(x_i,z_i) \in Z_{m}}\sum_{(x_j,z_j) \in Z_{n}}{d_{ij}}}{\lvert Z_{m} \rvert\lvert Z_{n} \rvert},\\
	d_{ij}& = \mathbb{I}[\lvert x_i-x_j \rvert+\lvert y_i-y_j \rvert\leq \epsilon].
    \end{split}
\end{equation}
where $\lvert Z_{m} \rvert$ is the element number, $\mathbb{I}[\cdot]$ is an indicative function, and $\epsilon$ is a hyper-parameter threshold. If the condition inside the brackets is satisfied, $\mathbb{I}[\cdot]$ equals 1; otherwise, it equals 0. Given the fixed number of regions $M$, the knowledge graph of each room exhibits a similar structure. To align the knowledge graphs across different rooms within the same kind of scene (\textit{e.g.}, living room, kitchen, bedroom, and bathroom), we employ the Kuhn-Munkres algorithm for pairwise matching. Subsequently, we merge the nodes and edges by averaging their values, ensuring a cohesive representation across the diverse rooms within the scene. It is important to note that traditional knowledge-graph-based navigators \cite{wu2019bayesian,yang2019visual,zhang2021hierarchical,zhang2023layout} construct nodes using discrete one-hot vectors and vote counting strategy, which fails to fully exploit the information contained in the natural language modality. As a result, these models exhibit limited generalization ability. In contrast, our knowledge graph employs continuous feature representations, enabling more powerful zero-shot generalization.

\noindent \textbf{Graph Adaptation.}\; Building precise representations of every specific scene is challenging, especially in new, unfamiliar environments with varying layouts. To address this, we dynamically update the node features of the initial knowledge graph based on real-time first-person observations. This process enables the initial knowledge graph with general prior information to evolve and adapt to the current environment. 
\begin{equation}
	\label{eq:update}
	\delta(V^{t})=\lambda I_{Z_C}f_{(x_t,z_t)}^T+(I-\lambda I_{Z_C}I_{Z_C}^T)\delta(V^{t-1}).
\end{equation}
where matrix $V^{0}\in \mathbb{R}^{M\times N}$ encapsulates the attributes of all nodes $V$ in the knowledge graph $\Omega = (V, E)$ at time stamp $t=0$, $I_{Z_C}\in \mathbb{R}^{M\times 1}$ describes the node index of the zone $Z_C$ where the agent is currently located, and $N$ is the feature length, which is 1024. The proximity between the current observation feature $f_{(x_t,z_t)}$ and the cluster centers can be computed, enabling the identification of the closest cluster center, which serves as the current region $Z_C$. Note that, the observed feature $f_{(x_t,z_t)}$ here is not required to traverse all views $(\theta_1, \theta_2)$ of the same location $(x, z)$. In addition, $\lambda$ represents a trainable parameter that governs the global impact of the currently observed feature $f_{(x_t,z_t)}$ on the knowledge graph. It is crucial to emphasize that these updates are episode-specific, as each episode within the scenes of the same category starts with the identical initial knowledge graph.

\subsection{Navigation Policy}
\label{sec:navigation}

The hierarchical navigation policy comprises high-level and low-level controllers. The high-level controller harnesses the power of the knowledge graph to effectively plan sub-goals for guiding the movement of agent across diverse regions. In parallel, the low-level controller leverages the integrated multimodal input to exhibit motion control.

\noindent \textbf{High-level Controller.}\; This paper defines the target zone $Z_T$ as the region in the initial knowledge graph with the highest probability of containing the target object $g$. The objective of the agent is to plan a path that starts from the current zone $Z_C$ and reaches the target zone $Z_T$ with the maximum connection probability, which is calculated as the product of all edges along the path $\prod_t e_t$. By doing so, real-time planning of sub-goal zones $Z_S$ becomes possible. Meanwhile, we leverage the power of GCN, a specialized neural network designed to process graph-structured data. GCN provides an effective framework for modeling and comprehending intricate relationships within knowledge graphs. The GCN output generates node-level representations while preserving the number of nodes and the length of node features. In Fig.\;\ref{fig:pipeline}, the output Graph Feature $f_{gra}\in \mathbb{R}^{1024}$ of the high-level controller refers to a specific node feature extracted from the GCN output, corresponding to the sub-goal zone. With the utilization of GCN, our approach enables the agent to make informed decisions based on the understanding of complex relationships within the knowledge graph. By incorporating GCN into our framework, we enhance the ability of agent to plan paths towards sub-goals in a more efficient and effective manner.

\noindent \textbf{Low-level Controller.}\; Similar to the traditional knowledge-graph-based navigators, AKGVP incorporates LSTM to model the low-level controller and learn the policy $\pi(a_t | s_t, g)$. The network takes a concatenation of Image Feature $f_{img}$, Object Feature $f_{obj}$, Graph Feature $f_{gra}$, and Previous Action $f_{act}$ as input. Among them, $f_{act}$ is the historical action of the last timestamp decision by agent, represented by a 6-dimensional one-hot vector. To optimize the performance of the low-level controller, we employ the A3C \cite{mnih2016asynchronous} algorithm within the framework of reinforcement learning. This algorithm enables effective training and policy optimization, empowering the agent to make informed decisions and navigate efficiently within the environment.

\section{Experiments}
\label{sec:exp}

\subsection{Experimental Setup}
\noindent \textbf{Datasets.}\; We evaluate the efficacy of our approach using AI2-THOR \cite{ai2thor} simulator, renowned for its provision of highly realistic 3D indoor scenes with near-photographic observations. The AI2-THOR dataset encompasses a diverse collection of 120 scenes across four categories: living room, kitchen, bedroom, and bathroom, featuring a wide range of 22 kinds of goal objects \{AlarmClock, Book, Bowl, CellPhone, Chair, CoffeeMachine, DeskLamp, FloorLamp, Fridge, GarbageCan, Kettle, Laptop, LightSwitch, Microwave, Pan, Plate, Pot, RemoteControl, Sink, StoveBurner, Television, Toaster\}. Each scene includes a minimum of 4 goal objects. Following the conventions set by prior research \cite{du2021vtnet,du2020learning,zhang2021hierarchical,zhang2023layout}, we allocate 20 rooms for training, 5 for validation, and 5 for testing purposes, ensuring a consistent and fair experimental setup. Furthermore, to validate the zero-shot capability of the navigators, we deliberately excluded the six goal objects \{Bowl, DeskLamp, Laptop, LightSwitch, Plate, StoveBurner\} from the training set. Consequently, the test set exclusively focuses on evaluating the performance with respect to these six specific goal objects. This allows us to assess the generalization ability of navigators in dealing with unseen objects.

\noindent \textbf{Settings.}\; In the framework setting, a reward system is introduced aimed at minimizing the trajectory length to the goal. Specifically, when the agent successfully reaches the goal object within a specified number of steps, it receives a significant positive reward of 5. Conversely, for each step taken towards the goal without reaching it, a small negative reward of -0.01 is imposed. For AI2-THOR-related settings, please refer to Sec.\;\ref{sec:objnav}. Adam optimizer is adopted to update the network parameters with a learning rate of $10^{-4}$. To train our models, a total of 6 million episodes is conducted, ensuring that each episode involves randomly selecting the starting position and goal for the agent.

\noindent \textbf{Evaluations.}\; We conducted a rigorous evaluation by conducting experiments in triplicate, and the reported results are presented as mean $\pm$ standard deviation. To assess the performance of our approach, we selected three evaluation metrics: Success Rate (SR), Success weighted by Path Length (SPL), and Distance to Goal (DTS). In the tables below, an upward arrow ($\uparrow$) indicates that a higher value is desirable, while a downward arrow ($\downarrow$) indicates that a lower value is preferred. SR evaluates the ability of agent to successfully locate the target object, while SPL takes into account both success rate and path length. DTS quantifies the distance between the agent and the goal at the end of each episode.

\subsection{Performance Analysis}
\begin{figure*}[t!]
	\centering
	\includegraphics[width=0.9\linewidth]{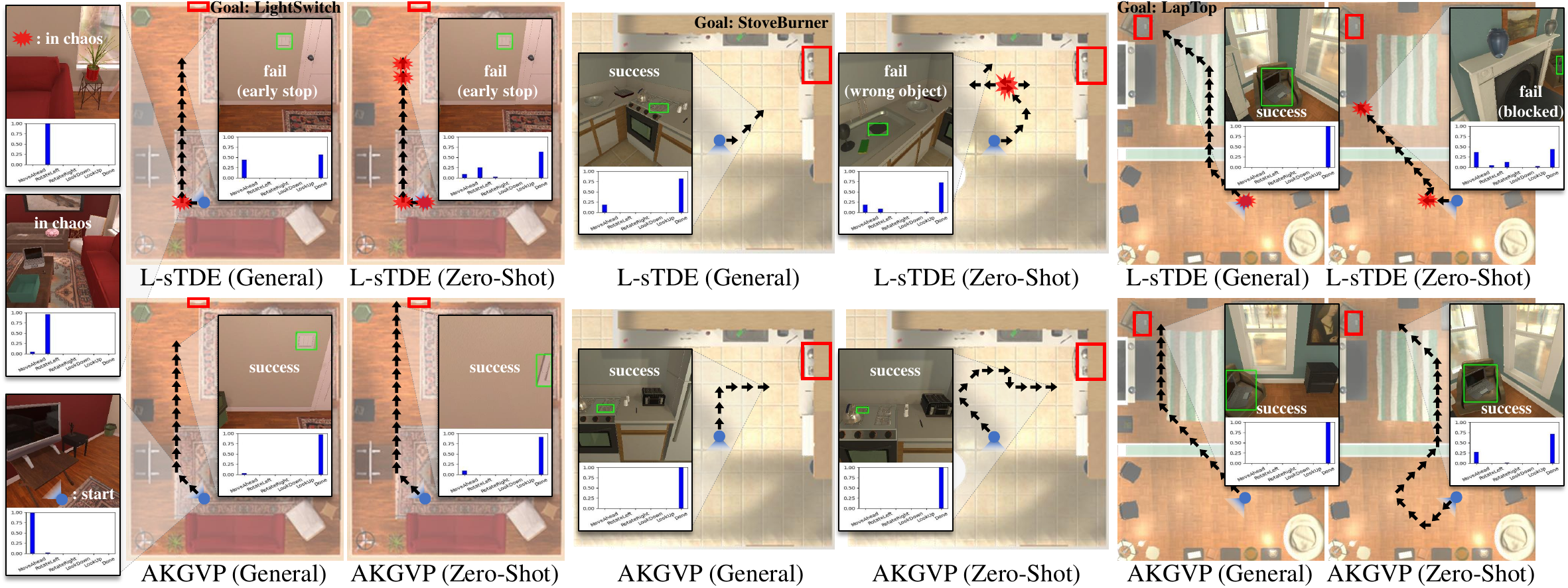}
    \vspace{-0.3cm}
	\caption{Qualitative results (zoom in for detailed viewing). The visualization of navigation results for the four navigators in three rooms is presented from left to right, along with the corresponding observations from the final frame of navigation. The red explosion icon denotes instances where the agent becomes disoriented and exhibits erratic behavior, such as spinning in circles, getting stuck by obstacles, or repetitive small-scale rotations. The blue circle represents the starting position, which is consistent across all four navigators. The histogram provides a visual representation of the action probabilities, highlighting the likelihood of the agent selecting each of the six actions. For additional instances of the comparison between these two methods, please refer to the accompanying video.}
	\label{fig:results}
	\vspace{-0.5cm}
\end{figure*}

To provide comprehensive evidence of the superiority of AKGVP, we conduct both qualitative and quantitative analyses, comparing it with the State-Of-The-Art (SOTA) navigators, \textit{e.g.}, HOZ \cite{zhang2021hierarchical} and L-sTDE \cite{zhang2023layout}. The quantitative results are presented in Tab.\;\ref{tab:general} and Tab.\;\ref{tab:zeroshot}, while the qualitative results are depicted in Fig.\;\ref{fig:results}. In Tab.\;\ref{tab:general} and Tab.\;\ref{tab:zeroshot}, AKGVP-Base refers to the algorithm described earlier, while AKGVP-CI, based on the AKGVP-Base version, incorporates the L-sTDE \cite{zhang2023layout} method by using the causal inference algorithm to eliminate the prediction bias brought by the experience, \textit{e.g.}, prior knowledge from the knowledge graph. These analyses offer a thorough evaluation and highlight the advantages of AKGVP in ObjNav tasks.

\noindent \textbf{General Navigation Performance.}\; As depicted in Tab.\;\ref{tab:general}, our proposed AKGVP method showcases outstanding performance in the regular ObjNav task when compared to SOTA navigators. Specifically, AKGVP-CI achieves the highest success rate (SR) of 76.78\% and the shortest distance to the goal (DTS) of 0.35m, surpassing the L-sTDE method by 2.59\% and 0.08m, respectively. Moreover, AKGVP-Base demonstrates the highest success weighted by path length (SPL) with a value of 40.66\%. It is worth noting that the SPL value of AKGVP-CI might not be optimal due to the agent being in closer proximity to the target when the navigation stops, resulting in additional steps. For evaluation purposes, an agent is deemed successful without discrimination if it is within a distance of 1.5m from the goal object. These results in Tab.\;\ref{tab:general} collectively underline the superior performance of our method, showcasing higher success rates and more efficient path planning in general ObjNav task, when compared to the baselines and other state-of-the-art approaches.

\begin{table}[t!]
	\centering
	\footnotesize
	\caption{General navigation performance comparison.}
	\vspace{-0.3cm}
	\label{tab:general}
	\setlength{\tabcolsep}{2.9mm}{
	\begin{tabular}{l|rrr}
		\toprule
		Navigator & SR$\uparrow$(\%) & SPL$\uparrow$(\%) & DTS$\downarrow$(m)  \\
		\midrule
		\midrule
		Random & 4.35\textcolor[rgb]{0.4,0.4,0.4}{\scriptsize{$\pm$1.69}} & 2.39\textcolor[rgb]{0.4,0.4,0.4}{\scriptsize{$\pm$1.37}} & 1.41\textcolor[rgb]{0.4,0.4,0.4}{\scriptsize{$\pm$0.01}} \\
		Baseline \cite{zhu2017target} & 58.76\textcolor[rgb]{0.4,0.4,0.4}{\scriptsize{$\pm$0.18}} & 34.48\textcolor[rgb]{0.4,0.4,0.4}{\scriptsize{$\pm$0.43}} & 0.68\textcolor[rgb]{0.4,0.4,0.4}{\scriptsize{$\pm$0.01}} \\
		SP \cite{yang2019visual} & 62.19\textcolor[rgb]{0.4,0.4,0.4}{\scriptsize{$\pm$0.67}} & 37.60\textcolor[rgb]{0.4,0.4,0.4}{\scriptsize{$\pm$0.35}} & 0.61\textcolor[rgb]{0.4,0.4,0.4}{\scriptsize{$\pm$0.02}} \\
		SAVN \cite{wortsman2019learning} & 63.27\textcolor[rgb]{0.4,0.4,0.4}{\scriptsize{$\pm$0.11}} & 38.20\textcolor[rgb]{0.4,0.4,0.4}{\scriptsize{$\pm$0.04}} & 0.56\textcolor[rgb]{0.4,0.4,0.4}{\scriptsize{$\pm$0.01}} \\
		EOTP \cite{mayo2021visual} & 65.61\textcolor[rgb]{0.4,0.4,0.4}{\scriptsize{$\pm$0.25}} & 38.93\textcolor[rgb]{0.4,0.4,0.4}{\scriptsize{$\pm$0.10}} & 0.55\textcolor[rgb]{0.4,0.4,0.4}{\scriptsize{$\pm$0.01}} \\
		ORG \cite{du2020learning} & 66.53\textcolor[rgb]{0.4,0.4,0.4}{\scriptsize{$\pm$0.29}} & 39.00\textcolor[rgb]{0.4,0.4,0.4}{\scriptsize{$\pm$0.34}} & 0.54\textcolor[rgb]{0.4,0.4,0.4}{\scriptsize{$\pm$0.01}} \\
		ORG+TPN \cite{du2020learning} & 68.60\textcolor[rgb]{0.4,0.4,0.4}{\scriptsize{$\pm$0.29}} & 39.40\textcolor[rgb]{0.4,0.4,0.4}{\scriptsize{$\pm$0.17}} & 0.54\textcolor[rgb]{0.4,0.4,0.4}{\scriptsize{$\pm$0.01}} \\
		VTNet \cite{du2021vtnet} & 70.10\textcolor[rgb]{0.4,0.4,0.4}{\scriptsize{$\pm$1.00}} & 39.60\textcolor[rgb]{0.4,0.4,0.4}{\scriptsize{$\pm$0.10}} & 0.52\textcolor[rgb]{0.4,0.4,0.4}{\scriptsize{$\pm$0.01}} \\
		HOZ \cite{zhang2021hierarchical} & 70.48\textcolor[rgb]{0.4,0.4,0.4}{\scriptsize{$\pm$0.54}} & 39.84\textcolor[rgb]{0.4,0.4,0.4}{\scriptsize{$\pm$0.21}} & 0.48\textcolor[rgb]{0.4,0.4,0.4}{\scriptsize{$\pm$0.02}} \\
		L-sTDE \cite{zhang2023layout} & 74.19\textcolor[rgb]{0.4,0.4,0.4}{\scriptsize{$\pm$0.60}} & 40.30\textcolor[rgb]{0.4,0.4,0.4}{\scriptsize{$\pm$0.27}} & 0.43\textcolor[rgb]{0.4,0.4,0.4}{\scriptsize{$\pm$0.01}} \\
		\midrule
		\midrule
		\textit{Ours} AKGVP \scriptsize{(Base)} &  73.63\textcolor[rgb]{0.4,0.4,0.4}{\scriptsize{$\pm$0.56}} & \textbf{40.66}\textcolor[rgb]{0.4,0.4,0.4}{\scriptsize{$\pm$0.22}}  & 0.44\textcolor[rgb]{0.4,0.4,0.4}{\scriptsize{$\pm$0.02}}   \\
		\textit{Ours} AKGVP \scriptsize{(CI)} & \textbf{76.78}\textcolor[rgb]{0.4,0.4,0.4}{\scriptsize{$\pm$0.51}} & 39.63\textcolor[rgb]{0.4,0.4,0.4}{\scriptsize{$\pm$0.32}}  & \textbf{0.35}\textcolor[rgb]{0.4,0.4,0.4}{\scriptsize{$\pm$0.01}}    \\
		\bottomrule
	\end{tabular}}
	\vspace{-0.2cm}
\end{table}

\begin{table}[t!]
	\centering
	\footnotesize
	\caption{Zero-shot navigation performance comparison.}
	\vspace{-0.3cm}
	\label{tab:zeroshot}
	\setlength{\tabcolsep}{2.9mm}{
		\begin{tabular}{l|rrr}
			\toprule
			Navigator & SR$\uparrow$(\%) & SPL$\uparrow$(\%) & DTS$\downarrow$(m)  \\
			\midrule
			\midrule
			Baseline \cite{zhu2017target} & 19.34\textcolor[rgb]{0.4,0.4,0.4}{\scriptsize{$\pm$0.28}} & 8.81\textcolor[rgb]{0.4,0.4,0.4}{\scriptsize{$\pm$0.41}} & 1.20\textcolor[rgb]{0.4,0.4,0.4}{\scriptsize{$\pm$0.01}} \\
			EmbCLIP \cite{khandelwal2022simple} & 44.14\textcolor[rgb]{0.4,0.4,0.4}{\scriptsize{$\pm$0.54}} & 15.76\textcolor[rgb]{0.4,0.4,0.4}{\scriptsize{$\pm$0.21}} & 0.79\textcolor[rgb]{0.4,0.4,0.4}{\scriptsize{$\pm$0.01}} \\
			HOZ \cite{zhang2021hierarchical} & 47.54\textcolor[rgb]{0.4,0.4,0.4}{\scriptsize{$\pm$0.53}} & 15.05\textcolor[rgb]{0.4,0.4,0.4}{\scriptsize{$\pm$0.35}} & 0.76\textcolor[rgb]{0.4,0.4,0.4}{\scriptsize{$\pm$0.01}} \\
			L-sTDE \cite{zhang2023layout} & 54.75\textcolor[rgb]{0.4,0.4,0.4}{\scriptsize{$\pm$0.60}} & 16.64\textcolor[rgb]{0.4,0.4,0.4}{\scriptsize{$\pm$0.31}} & 0.62\textcolor[rgb]{0.4,0.4,0.4}{\scriptsize{$\pm$0.02}} \\
			\midrule
			\midrule
			\textit{Ours} AKGVP \scriptsize{(Base)} & 51.80\textcolor[rgb]{0.4,0.4,0.4}{\scriptsize{$\pm$0.51}} & 17.47\textcolor[rgb]{0.4,0.4,0.4}{\scriptsize{$\pm$0.28}} & 0.69\textcolor[rgb]{0.4,0.4,0.4}{\scriptsize{$\pm$0.01}}  \\
			\textit{Ours} AKGVP \scriptsize{(CI)} & \textbf{69.51}\textcolor[rgb]{0.4,0.4,0.4}{\scriptsize{$\pm$0.47}} & \textbf{28.86}\textcolor[rgb]{0.4,0.4,0.4}{\scriptsize{$\pm$0.25}} & \textbf{0.41}\textcolor[rgb]{0.4,0.4,0.4}{\scriptsize{$\pm$0.02}} \\
			\bottomrule
	\end{tabular}}
	\vspace{-0.4cm}
\end{table}

\noindent \textbf{Zero-shot Navigation Performance.}\; In comparison to the results presented in Tab.\;\ref{tab:general}, the zero-shot navigation performance demonstrated in Tab.\;\ref{tab:zeroshot} is particularly remarkable. The AKGVP-CI method demonstrates robust generalization capabilities even when encountering unknown objects. While there is a slight decline in the performance of AKGVP-CI compared to the general navigation scenario, with a decrease in SR by 7.27\%, an increase in SPL by 10.77\%, and a slight increase in DTS by 0.06m, the Baseline and SOTA navigators experience more significant setbacks with deteriorated performance indicators. Additionally, the zero-shot navigator EmbCLIP, which solely relies on the CLIP image encoder without utilizing the CLIP language encoder and high-level controller, also underperforms. These results highlight the importance of the continuous knowledge graph architecture, as well as multimodal feature alignment, in achieving superior performance in zero-shot navigation tasks.

\noindent \textbf{Visualization and Qualitative Study.}\; As depicted in Fig.\;\ref{fig:results}, AKGVP-CI exhibits robust generalization capabilities in both general navigation and zero-shot navigation tasks. While the state-of-the-art navigator L-sTDE excels in general navigation, it frequently encounters confusion in zero-shot navigation, \textit{e.g.}, premature stops, incorrect target identification, and progress blocked by obstacles. Furthermore, AKGVP-CI demonstrates more precise action decision-making, ensuring smoother and more efficient navigation trajectories.

\subsection{Ablation Study}
In this section, we conduct three types of ablation experiments to investigate the impact of different pre-training settings (Tab.\;\ref{tab:pretraining}) and model components (Tab.\;\ref{tab:components}) on the overall performance of our model. The objective is to provide insights into the key factors influencing performance.

\begin{table}[t!]
	\footnotesize
	\centering
	\caption{Ablation study on pre-training settings.}
	\vspace{-0.3cm}
	\label{tab:pretraining}
	\setlength{\tabcolsep}{1.5mm}{
		\begin{tabular}{lc|rrr}
			\toprule
			Pre-training   & Frozen Parameters & SR$\uparrow$(\%) & SPL$\uparrow$(\%) & DTS$\downarrow$(m) \\
			\midrule
			ImageNet \cite{deng2009imagenet} & \cmark & 73.66 & 37.21 & 0.46 \\
			ImageNet \cite{deng2009imagenet} & \xmark & 74.12 & 38.31 & 0.43 \\
			CLIP \cite{radford2021learning} & \cmark & \textbf{76.78} & 39.63 & \textbf{0.35} \\
			CLIP \cite{radford2021learning} & \xmark & 75.86 & \textbf{40.97} & 0.40 \\
			\bottomrule
	\end{tabular}}
	\vspace{-0.2cm}
\end{table}

\begin{table}[t!]
	\footnotesize
	\centering
	\caption{Ablation study on model component decomposition.}
	\vspace{-0.3cm}
	\label{tab:components}
	\setlength{\tabcolsep}{2.2mm}{
		\begin{tabular}{cccc|rrr}
			\toprule
			$f_{img}$   & $f_{obj}$  & $f_{gra}$  & $f_{act}$ & SR$\uparrow$(\%) & SPL$\uparrow$(\%) & DTS$\downarrow$(m) \\
			\midrule
			\xmark & \cmark & \xmark & \xmark & 9.47 & 1.45 & 1.22 \\
			\cmark & \xmark & \xmark & \xmark & 16.80 & 2.81 & 1.37 \\
			\cmark & \cmark & \xmark & \xmark & 69.57 & 37.58 & 0.48 \\
			\cmark & \cmark & \cmark & \xmark & 73.63 & 38.82 & 0.38 \\
			\cmark & \cmark & \cmark & \cmark & \textbf{76.78} & \textbf{39.63} & \textbf{0.35} \\
			\bottomrule
	\end{tabular}}
	\vspace{-0.5cm}
\end{table}

\noindent \textbf{Form of Pre-training.}\; The purpose of this ablation experiment is to assess the effectiveness of visual language pre-training. While keeping the model structure unchanged, we replace the visual encoder in AKGVP with the ResNet-50 \cite{he2016deep} pre-trained on the ImageNet \cite{deng2009imagenet} dataset, and remove the language encoder, thereby reverting to the one-hot encoding approach by the traditional navigator. The results are recorded in Tab.\;\ref{tab:pretraining}. It is evident that the algorithm without multimodal pre-training alignment exhibits an average decrease of 2.28\% in SR, 2.54\% in SPL, and an increase of 0.06m in DTS compared to the aligned algorithm. This indicates the importance and effectiveness of multimodal pre-training for achieving superior navigation performance.

\noindent \textbf{Frozen Parameters.}\; To assess the impact of frozen parameters on navigation performance, an experiment is conducted that investigated whether to freeze the parameters of the pre-training model in Tab.\;\ref{tab:pretraining}. When the pre-trained model lacks multimodal alignment, unfreezing the parameters leads to improved navigation performance. Specifically, the SR value increases by 0.46\%, the SPL value increases by 1.10\%, and the DTS decreases by 0.03m. In contrast, navigators that have undergone multimodal alignment show no significant performance gains when unfreezing the encoder parameters. This observation indicates that the contribution of unfreezing the parameters to training the encoder is limited.

\noindent \textbf{Model Components.}\; To investigate the impact of the four inputs (Image Feature $f_{img}$, Object Feature $f_{obj}$, Graph Feature $f_{gra}$, Previous Action $f_{act}$) on the decision-making process of the low-level controller, we conducted a series of ablation experiments by dissecting the model structure, as presented in Tab.\;\ref{tab:components}. The results reveal that $f_{img}$ and $f_{obj}$ exert the most significant influence, with the removal of either input resulting in a substantial decline in navigator performance. On the other hand, the inclusion of sub-goal information $f_{gra}$ provided by the high-level controller and historical action information $f_{act}$ enhances path planning capabilities, leading to slight improvements in navigation performance.

\section{Conclusion}
The contributions of this research lie in the development of the Aligning Knowledge Graph with Visual Perception (AKGVP) method for the ObjNav task, which addresses the misalignment between discrete scene features and first-person visual observations through continuous knowledge graph modeling and visual-language pre-training. Furthermore, we provide comprehensive evaluations of AKGVP, showcasing its superior performance and efficiency in both general and zero-shot object-goal navigation tasks. By aligning language description with visual perception, AKGVP holds promise for advancing the field of embodied intelligence, enabling more accurate and effective navigation in dynamic environments.

\noindent \textbf{Acknowledgment.}\; This work is supported by the National Key R\&D Program of China (2022YFB4501600) and the National Natural Science Foundation of China (No. U21A20488).


\bibliographystyle{IEEEtran}
\bibliography{icra}

\end{document}